\documentclass[11pt]{article}
\usepackage{nodalida2021}
\usepackage{times}
\usepackage{url}
\usepackage{latexsym}
\usepackage{etaremune}
\usepackage{multirow}
\usepackage{graphicx}
\usepackage{arydshln}
\usepackage{longtable}

\usepackage{xcolor}
\aclfinalcopy % Uncomment this line for the final submission
\title{Finnish Paraphrase Corpus}

\author{Jenna Kanerva, Filip Ginter, Li-Hsin Chang, Iiro Rastas, Valtteri Skantsi,\\ \textbf{Jemina Kilpel{\"a}inen, Hanna-Mari Kupari, Jenna Saarni, Maija Sev{\'o}n, and Otto Tarkka}\\
  TurkuNLP group \\
  Department of Computing \\
  Faculty of Technology \\
  University of Turku, Finland\\
  {\tt jmnybl@utu.fi}}

\date{}

\begin{document}
\maketitle
\begin{abstract}
  In this paper, we introduce the first fully manually annotated paraphrase corpus for Finnish containing 53,572 paraphrase pairs harvested from alternative subtitles and news headings. Out of all paraphrase pairs in our corpus 98\% are manually classified to be paraphrases at least in their given context, if not in all contexts. Additionally, we establish a manual candidate selection method and demonstrate its feasibility in high quality paraphrase selection in terms of both cost and quality.
\end{abstract}

\section{Introduction}
The powerful language models that have recently become available in NLP have also resulted in a distinct shift towards more meaning-oriented tasks for model fine-tuning and evaluation. The most typical example is entailment detection, with the paraphrase task raising in interest recently.  
Paraphrases, texts that express the same meaning with differing words \cite{bhagat-hovy-2013-squibs}, are --- already by their very definition --- a suitable target to induce and evaluate models' ability to represent meaning. Paraphrase detection and generation has numerous direct applications in NLP \cite{madnani-dorr-2010-review}, among others in question answering \cite{soni-roberts-2019-QA}, plagiarism detection \cite{Altheneyan2020plagDectionReview}, and machine translation \cite{mehdizadeh-seraj-etal-2015-improving}.

Research in paraphrase naturally depends on the availability of datasets for the task. We will review these in more detail in Section~\ref{sec:related-work}, nevertheless, barring few exceptions, paraphrase corpora are typically large and gathered automatically using one of several possible heuristics. Typically a comparatively small section of the corpus is manually classified to serve as a test set for method development. The heuristics used to gather and filter the corpora naturally introduce a bias to the corpora which, as we will show later in this paper, demonstrates itself as a tendency towards short examples with a relatively high lexical overlap. Addressing this bias to the extent possible, and providing a corpus with longer, lexically more diverse paraphrases is one of the motivations for our work. The other motivation is to cater for the needs of Finnish NLP, and improve the availability of high-quality, manually annotated paraphrase data specifically for the Finnish language.

In this paper, we therefore aim for the following contributions: Firstly, we establish and test a fully manual procedure for paraphrase candidate selection with the aim of avoiding a selection bias towards short, lexically overlapping candidates.
Secondly, we release the first fully manually annotated paraphrase corpus of Finnish, sufficiently large for model training. The number of manually annotated examples makes the released dataset one of the largest, if not the largest manually annotated paraphrase corpus for any language. And thirdly, we report the experiences, tools, and baseline results on this new dataset, hopefully allowing other language NLP communities to assess the potential of developing a similar corpus for other languages.

\section{Related Work}
\label{sec:related-work}

\begin{table*}[]
  \centering
  \begin{tabular}{llccc}
    \hline \textbf{Corpus} & \textbf{Data source} & \textbf{Size autom.} & \textbf{Size manual} &  \textbf{Labels} \\\hline
    \textbf{English} & & & & \\
    MRPC & Online news & --- & 5,801 & 0/1 \\
    TUC & News tweets & --- & 52K & 0/1 \\
    ParaSCI & Scientific papers & 350K & --- & 1-5 \\
    PARADE & flashcards (computer sci.) & --- & 10K & 0-3 \\
    QQP & Quora & 404K & --- & 0/1 \\
    \hline \textbf{Finnish} & & & \\
    Opusparcus & OpenSubtitles & 480K* & 3,703 & 1-4 \\
    TaPaCo & Tatoeba crowdsourcing & 12K & --- & --- \\\hline
  \end{tabular}
  \caption{Summary of available paraphrase corpora of naturally occurring sentential paraphrases. The corpora sizes include the total amount of pairs in the corpus (i.e. also those labeled as non-paraphrases), thus the actual number of good paraphrases depend on the class distribution of each corpus. *The highest quality cutpoint estimated by the authors.}
  \label{tab:related-corpora}
\end{table*}

% method summary:
% MRPC: heuristics + classifier + manual annotation
% QQP: merge/no-merge labels in the forum
% TUC: heuristics + manual annotation
% PARADE: heuristics + clustering (candidates taken only from the same cluster) + manual annotation
% ParaSCI: heuristics + classifier
% Opusoarcus: language pivoting
% TaPaCo: language pivoting

% PIT https://www.aclweb.org/anthology/S15-2001.pdf ???

% heuristics + filtering
%% @LI - MRPC - what heuristics, roughly?
Statistics of the different paraphrase corpora most relevant to our work are summarized in Table~\ref{tab:related-corpora}. For English, the \textbf{Microsoft Research Paraphrase Corpus (MRPC)}~\cite{dolan2005MSRP} is extracted from an online news collection by applying heuristics to recognize candidate document pairs and candidate sentences from the documents. Paraphrase candidates are subsequently filtered using a classifier, before the final manual binary annotation (paraphrase or not). In the \textbf{Twitter URL Corpus (TUC)} \cite{lan-etal-2017twitterUrlCorpus}, paraphrase candidates are identified by recognizing shared URLs in news related tweets. All candidates are manually binary-labeled. \textbf{ParaSCI} \cite{Dong2021ParaSCI} is created by collecting paraphrase candidates from ACL and arXiv papers using heuristics based on term definitions, citation information as well as sentence embedding similarity. The extracted candidates are automatically filtered, but no manually annotated data is available. \textbf{PARADE}~\cite{he-etal-2020-parade} is created by collecting online user-generated flashcards for computer science related concepts. All definitions for the same term are first clustered, and paraphrase candidates are extracted only among a cluster to reduce noise in candidate selection. All extracted candidates are manually annotated using a scheme with four labels. \textbf{Quora Question Pairs (QQP)}\footnote{\url{data.quora.com/First-Quora-Dataset-\Release-Question-Pairs}} contains question headings from the forum with binary labels into duplicate-or-not questions. The QQP dataset is larger than other datasets, however, although including human-produced labels, the labeling is not originally designed for paraphrasing and the dataset providers warn about labeling not guaranteed to be perfect.

% language pivoting
Another common approach for automatic paraphrase identification is through language pivoting using multilingual parallel datasets. Here sentence alignments are used to recognize whether two different surface realizations share an identical or near-identical translation, assuming that the identical translation likely implies a paraphrase. There are two different multilingual paraphrase datasets automatically extracted using language pivoting, Opusparcus~\cite{CREUTZ18.131} and TaPaCo~\cite{Scherrer2020TaPaCo}, both including a Finnish subsection. \textbf{Opusparcus} consists of candidate paraphrases automatically extracted from the alternative translations of movie and TV show subtitles after automatic sentence alignment. While the candidate paraphrases are automatically extracted, a small subset of few thousand paraphrase pairs for each language is manually annotated.
\textbf{TaPaCo} contains candidate paraphrases automatically extracted from the Tatoeba dataset\footnote{\url{https://tatoeba.org/eng/}}, which is a multilingual crowdsourced database of sentences and their translations. Like Opusparcus, TaPaCo is based on language pivoting, where all alternative translations for a same statement are collected. However, unlike most other corpora, the candidate paraphrases are grouped into `sets' instead of pairs, and all sentences in a set are considered equivalent in meaning. TaPaCo does not include any manual validation.

\section{Text Selection}
\label{sec:data-selection}

As discussed previously, we elect to rely on fully manual candidate extraction as a measure against any bias introduced through heuristic candidate selection methods. In order to obtain sufficiently many paraphrases for the person-months spent, the text sources need to be paraphrase-rich, i.e. have a high probability for naturally occurring paraphrases. Such text sources include for example news headings and articles reporting on the same news, alternative translations of the same source material, different student essays and exam answers for the same assignment, and related questions with their replies in discussion fora, where one can assume different writers using distinct words to state similar meaning. For this first version of the corpus, we use two different text sources: alternative Finnish subtitles for the same movies or TV episodes, and headings from news articles discussing the same event in two different Finnish news sites.

\subsection{Alternative Subtitles}

OpenSubtitles\footnote{\url{http://www.opensubtitles.org}} distributes an extensive collection of user generated subtitles for different movies and TV episodes. These subtitles are available in multiple languages, but surprisingly often the same movie or episode have versions in a single language, originating from different sources. This gives an opportunity to exploit the natural variation produced by independent translators, and by comparing two different subtitles for a single movie or episode, there is a high likelihood of finding naturally occurring paraphrases.

% how could we call movie/episode with one term?

From the database dump of OpenSubtitles2018 obtained through OPUS~\cite{tiedemann-2012-OPUS}, we selected all movies and TV episodes with at least two Finnish subtitle versions. In case more versions are available, the two most lexically differing are selected for paraphrase extraction. To filter out subtitle pairs with low density of interesting paraphrase candidates, pairs with too high or too low cosine similarity of TF-IDF weighted document vectors are discarded. High similarity usually reflects identical subtitles with minor formatting differences, while low similarity is typically caused by incorrect identifiers in the source data. The two selected subtitle versions are then roughly aligned using the timestamps, and divided into segments of 15 minutes. For every movie/episode, the annotators are assigned one random such segment, the two versions presented side-by-side in a custom tool, allowing for fast selection of paraphrase candidates.

In total, we were able to obtain at least one pair of aligned subtitle versions for 1,700 unique movies and TV series. While for each unique movie only one pair of aligned subtitles is selected for annotation, TV series comprise different episodes, dealing with the same plot and characters, and therefore overlapping in language. After an initial annotation period, we noticed a topic bias towards a limited number of TV series with a large number of episodes, and decided to limit the number of annotated episodes to 10 per each TV series in all subsequent annotation. In total, close to 3,000 different movies/episodes are used for manual paraphrase candidate extraction, each including exactly one pair of aligned subtitles.

\subsection{News Headings}

We have downloaded news articles through open RSS feeds of different Finnish news sites during 2017--2021, resulting in a substantial collection of news from numerous complementary sources. For this present work, we narrow the data down to two sources: the Finnish Broadcasting Company (YLE) and Helsingin Sanomat (HS, English translation: Helsinki News). We align the news using a 7-day sliding window on time of publication, combined with cosine similarity of TF-IDF-weighted document vectors induced on the article body, obtaining article pairs likely reporting on the same event. We use the article headings as paraphrase candidates, striving to select maximally dissimilar headings of maximally similar articles as the most promising candidates for non-trivial paraphrases. In practice, we used a simple grid search and human judgement to establish the most promising region of article body and heading similarity values.

\section{Paraphrase Annotation}

The paraphrase annotation is comprised of multiple annotation steps, including candidate selection as described above, manual classification of candidates based on an annotation scheme, as well as the possibility of rewriting partial paraphrases into full paraphrases. Next, we will discuss the different paraphrase types represented in our annotation scheme, and afterwards the annotation workflow is discussed in a more detailed fashion.

\subsection{Annotation Scheme}
\label{sec:scheme}

Instead of a simple yes/no (\emph{equivalent} or \emph{not equivalent}) as in MRPC \cite{dolan2005MSRP} or 1--4 scale (\emph{bad}, \emph{mostly bad}, \emph{mostly good} and \emph{good}) as in Opusparcus \cite{CREUTZ18.131}, our annotation scheme is adapted to capture the level of paraphrasability in a more detailed fashion. Our annotation scheme uses the base scale 1--4 similar to other paraphrase corpora, enriched with additional subcategories (flags) for distinguishing different types of paraphrases which would otherwise fall from the label 4 (\emph{good}) into label 3 (\emph{mostly good}).

An example for each of the categories discussed below is shown in Table~\ref{tab:annotations} (English translations available in Appendix \ref{sec:appendix}). Each candidate pair is first evaluated in terms of the base scale numbered from 1 to 4, where:

\paragraph{Label 4} is a full (perfect) paraphrase in all reasonably imaginable contexts, meaning one can always be replaced with the other without changing the meaning. This ability to substitute one for the other in any context is the primary test for \emph{label 4} used in the annotation.
\paragraph{Label 3} is a context dependent paraphrase, where the meaning of the two statements is the same in the present context, but not necessarily in other contexts.
\paragraph{Label 2} is related but not a paraphrase, where there is a clear relation between the two statements, yet they cannot be considered paraphrases.
\paragraph{Label 1} is unrelated, there being no reasonable relation between the two statements, most likely a false positive in candidate selection.

\paragraph{}If labeling a candidate pair is not possible for a reason, or giving a label would not serve the desired purpose (e.g. wrong language or identical statements), the example can be skipped with the \emph{label x}.

With the base labels alone, a great number of candidate paraphrases would fail the substitution test for \textit{label 4} and be classified \emph{label 3}. This is especially so for longer text segments which are less likely to express strictly the same meaning. In order to avoid populating the \emph{label 3} category with a very diverse set of paraphrases, we opt to introduce flags for finer sub-categorization and thus support a broader range of downstream applications of the corpus. These flags are always attached to \emph{label 4} (subcategories of full paraphrases), meaning the paraphrases are not fully interchangeable due to the specified reason, but, crucially, are context-independent, unlike \emph{label 3}. The possible flags are:

\paragraph{Subsumption ($>$ or $<$)} where one of the statements is more detailed and the other more general. The relation of the pair is therefore directional, where the more detailed statement can be replaced with the more general one in all contexts, but not the other way around. The two common cases are one statement having additional minor details the other omits, and one statement being ambiguous while the other not. If there is a justification for crossing directionality (one statement being more detailed in one aspect while the other in another aspect), the pair falls into \emph{label 3} as the directional replacement test does not hold anymore.

\paragraph{Style (s)} for tone or register difference in cases where the meaning of the two statements is the same, but the statements differ in tone or register such that in certain situations, they would not be interchangeable. For example, if one statement uses pejorative language or profanities, while the other is neutral, or one is clearly colloquial language while the other is formal. The style flag also includes differences in the level of politeness, uncertainty, and strength of the statements.

\paragraph{Minor deviation (i)} marks in most cases minimal differences in meaning (typically "this" vs. "that") as well as easily traceable differences in grammatical number, person, tense or such. Some applications might consider these as \emph{label 4} for all practical purposes (e.g.\ information retrieval), while others should regard these as \emph{label 2} (e.g.\ automatic rephrasing). 

% Search engine:
% Milloin minun kannattaisi soittaa poliisille?
% Koska meidän pitäisi soittaa poliisille?

\paragraph{}The flags are independent of each other and can be combined in the annotation.

\begin{table*}[]
    \centering
    \begin{tabular}{lll}\hline
       Label  & Statement 1 & Statement 2  \\\hline
         4 & Tyrmistyttävän lapsellista! & Pöyristyttävän kypsymätöntä! \\
         4s & Olen työskennellyt lounaan ajan. & Tein töitä koko ruokiksen. \\
         4i & Teitäpä onnisti. & Oletpa onnekas. \\
         4$>$ & Tein lujasti töitä niiden rahojen eteen. & Paiskin kovasti töitä. \\
         4$<$s & Sä ruletat! Anna mennä! & Sinä olet paras, Tähkä! Anna mennä! \\
         4is & Sä pöllit meidän kasvin! & Varastit meidän kasvit! \\
         3 & Aion tehdä kokeen. & Aion testata sitä. \\
         2 & Tappion kokenut Väyrynen katosi Helsingin yöhön. & Väyrynen putoamassa eduskunnasta. \\\hline
          & Rewrites & \\\hline
         Orig & Voinko palata tehtäviini? & Saanko jatkaa? \\
         Rew    & \emph{Voinko palata tehtäviini?} &\emph{Saanko jatkaa tehtäviäni?} \\

%         4$>$i & Mukava, että voit käsitellä seikkaa noin kevyesti. & Hyvä, että otatte tämän noin kevyesti. \\
%         4$<$is & Kuka toi tuon hohdokkaan hemmon? & Kuka toi tämän kiiltävän miehen? \\
%         1 & Jotkut ovat hyviä, toiset pahoja, jotkut erityisen. & Alkukantaisia, kehittyneitä. \\\hline
         
    \end{tabular}
    \caption{Example paraphrase pairs annotated with different labels and flags (English translations available in Appendix \ref{sec:appendix}).}
    \label{tab:annotations}
\end{table*}

\subsection{Annotation Workflow}

Given two aligned documents as described in Section~\ref{sec:data-selection}, an annotator first extracts all candidate paraphrases. These can be anything between a short phrase and several sentences long, typically being about a sentence long. The annotators are encouraged to select as long continuous statements as possible, nevertheless at the same time avoiding a bias towards subsumption flag by over-extending one of the candidates. The candidate paraphrases are subsequently transferred into a classification annotation tool. In case of news headings, where the candidates are extracted automatically, the candidates are introduced directly in the classification tool without any manual extraction step.

In the classification tool, the annotator assigns a label for each candidate. The candidate paraphrases are shown one pair at a time, and for each pair the document context is available.

In addition to assigning a label and optional flags for a candidate pair, the classification tool provides an option to rewrite the statements if the classification is anything else than \emph{label 4} without any flags. The annotators are instructed to rewrite the candidates in cases, where a simple fix, for example word or phrase deletion, addition or replacement with a synonym or changing an inflection, can be easily constructed. Rewrites must be such that the annotated label for the rewritten example is \emph{4}. In cases where the rewrite would require more complicated changes or would take too much time, the annotators are instructed to move on to the next candidate pair. One rewrite done during the data annotation is illustrated in Table~\ref{tab:annotations}. 

The annotators can mark unsure, difficult or otherwise interesting cases for later discussion in daily annotation meetings. The annotators also communicate online, for instance seeking a quick validation for a particular decision. The work is further supported by a jointly produced 17-page annotation manual, which is revised and extended regularly.

\section{Corpus Statistics and Evaluation}

\begin{figure}
    \centering
    \includegraphics[width=0.5\textwidth]{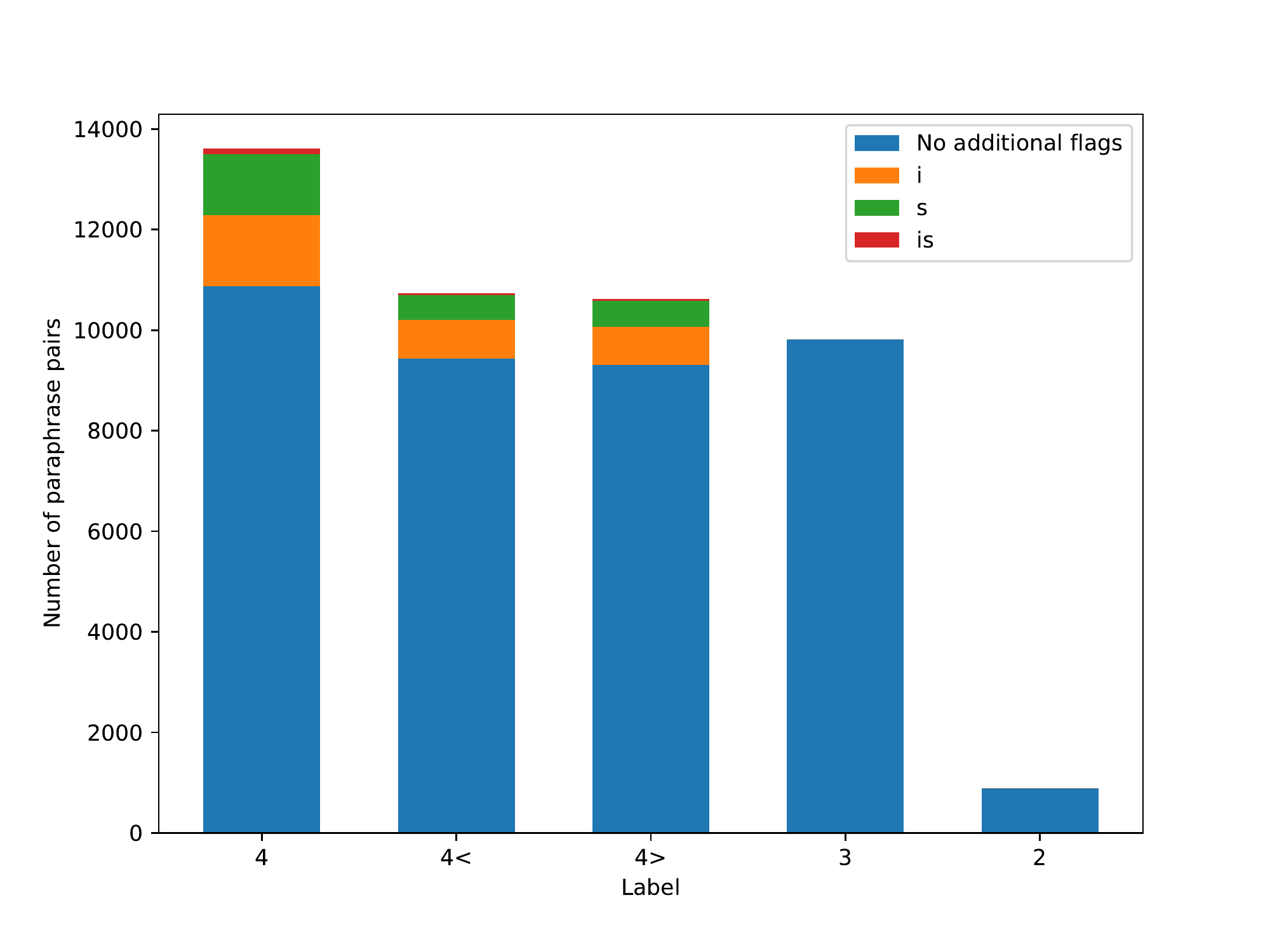}
    \caption{Labels distribution in our corpus.}
    \label{fig:label-distribution}
\end{figure}

The released corpus includes 45,663 naturally occurring paraphrases with additional 7,909 rewrites, resulting in the total size of 53,572 paraphrase pairs. Basic data statistics are summarized in Table~\ref{tab:stats}, and label distribution in Figure~\ref{fig:label-distribution}. Notably, the amount of candidate pairs labeled as not paraphrases (labels 1 or 2 in our scheme) is almost non-existent, owing to the manual candidate selection step in subtitles data from which the vast majority of the corpus data originates. Only 5.6\% of paraphrase pairs in the corpus originate from the automated candidate selection from news data. The amount of candidates labeled with label 1 or label x is insignificantly small, therefore we decided to discard these from the final corpus.

\begin{table}[]
    \centering
    \begin{tabular}{l|rr|r}
        \hline
        Section & Examples & Rewrites & Total  \\\hline
         Train  & 36,600   & 6,239    &  42,839 \\
         Devel  & 4,474    & 884      &  5,358  \\
         Test   & 4,589    & 786      &  5,375  \\\hline
         Total  & 45,663   & 7,909    & \textbf{53,572}   \\
    \end{tabular}
    \caption{Data sizes in our corpus.}
    \label{tab:stats}
\end{table}

\begin{figure}

\includegraphics[width=0.5\textwidth]{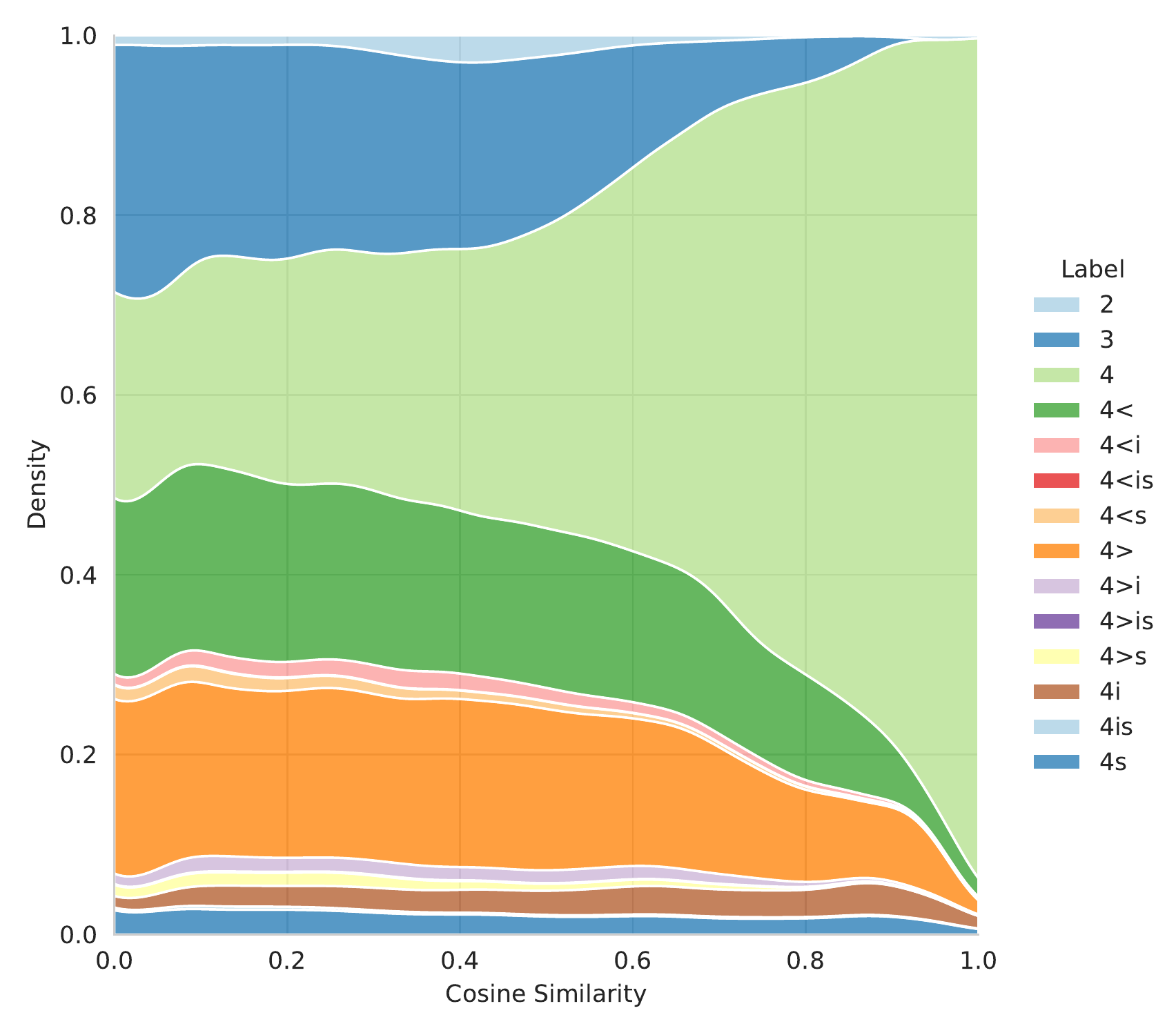}

\caption{Density of different labels in the training set conditioned on cosine similarity of the paraphrase pairs.}
\label{fig:label_cde}
\end{figure}

In Figure~\ref{fig:label_cde} we measure the density of different label combinations in the training set conditioned on cosine similarity of paraphrase pairs based on TF-IDF weighted character n-grams of lengths 2--4. Up to cosine similarity of 0.5 the most common labels are evenly represented, while the prevalence of label 4 increases throughout the range and dominates the sparsely populated range of similarities over 0.8.

\subsection{Annotation Quality}

After the initial annotator training phase most of the annotation work is carried out as single annotation. In order to monitor annotation consistency, double annotation batches are assigned regularly. In double annotation, one annotator first extracts the candidate paraphrases from the aligned documents, but later on these candidates are assigned to two different annotators, who annotate the labels for these independently from each other. Next, these two individual annotations are merged and conflicting labels are resolved together with the whole annotation team in a meeting. These consensus annotations constitute a gold standard against which individual annotators can be measured.

% double annotated examples: 1175
% total examples in the corpus : 45663

% agreements

% Full labels:
% Correct: 1727
% Total: 2513
% Accuracy: 0.6872264226024671
% Weighted averaged Cohen's Kappa: 0.41
% All annotations versus gold standard kappa: 0.62

% Base label and directional arrow:
% Correct: 1831
% Total: 2513
% Accuracy: 0.7286112216474333
% Weighted averaged Cohen's Kappa: 0.45
% All annotations versus gold standard kappa: 0.65

% Base label only:
% Correct: 2086
% Total: 2513
% Accuracy: 0.83008356545961
% Weighted averaged Cohen's Kappa: 0.46
% All annotations versus gold standard kappa: 0.65

A total of 1,175 examples are double annotated (2.5\% of the data\footnote{During the initial annotator training double annotation was used extensively; this annotator training data is not included in the released corpus.}). Most of these are annotated by exactly two annotators, however, some examples may include annotations from more than two annotators, and thus the total amount of individual annotations for which the gold standard label exists is 2,513. We measure the agreement of individually annotated examples against the gold standard annotations in terms of accuracy, i.e. the proportion of individually annotated examples with correctly assigned label.

The overall accuracy is 68.7\% when the base label (\emph{labels 1--4}) as well as all additional flags are taken into consideration. When discarding the least common flags \emph{s} and \emph{i} and evaluating only base labels and directional subsumption flags, the overall accuracy is 72.9\%. To compare the observed agreement to previous studies on paraphrase annotation, the Opusparcus annotation agreement is approximately 64\% on Finnish development set and 67\% on test set (calculated from numbers in Table 4 and Table 5 in \citet{CREUTZ18.131}). The Opusparcus uses an annotation scheme with four labels, similar to our base label scheme. In MRPC, the reported agreement score is 84\% on a binary paraphrase-or-not scheme. While direct comparison is difficult due to the different annotation schemes and label distributions, the figures show that the observed agreement seem to be roughly within the same range with agreement numbers seen in related works.

In addition to agreement accuracy, we calculate two versions of Cohen's kappa, a metric for inter-annotator agreement taking into account the possibility of agreement occurring by chance. First we measure kappa agreement of all individual annotations against the gold standard, an approach typical in paraphrase literature. This kappa is 0.62, indicating substantial agreement. Additionally, we measure the Cohen's kappa between each pair of annotators. The weighted average kappa over all annotator pairs is 0.41 indicating moderate agreement. Both are measured on full labels. When evaluating only on base labels and directional subsumption flags, these kappa scores are 0.65 and 0.45, respectively.

% compare opusparcus agreement ~64% (accuracy based on Table 4 "fi", labels 1–4), 67% (accuracy based on Table 5 "fi", labels 1–4)
% compare MSRP agreement 84%, kappa 0.62 (labels yes/no)

% <0 No agreement
%  0—.20 Slight
% .21—.40 Fair
% .41—.60 Moderate
% .61—.80 Substantial
% .81–1.0 Perfect

\subsection{Corpus Comparison}

\begin{figure*}

\includegraphics[width=0.33\textwidth]{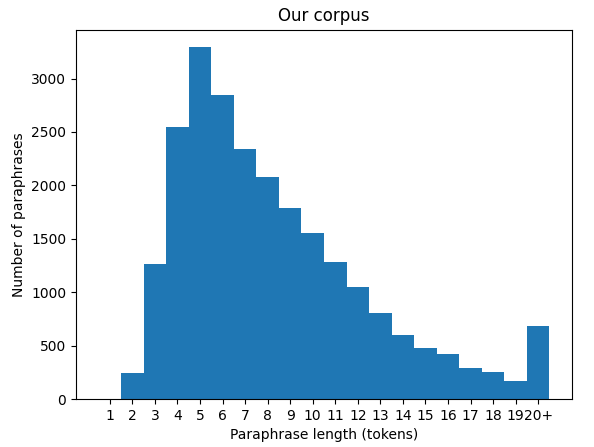}\hfill \includegraphics[width=0.33\textwidth]{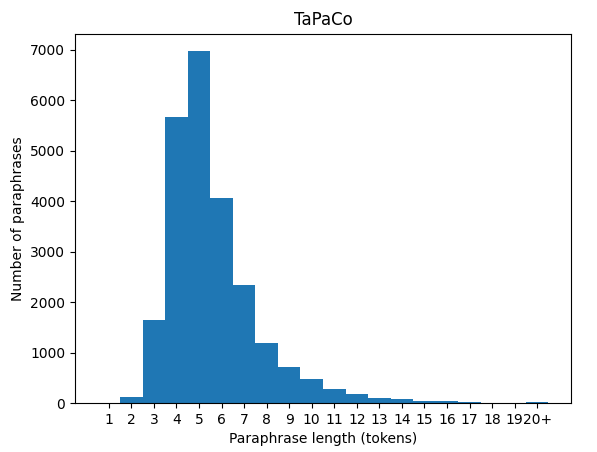}\hfill \includegraphics[width=0.33\textwidth]{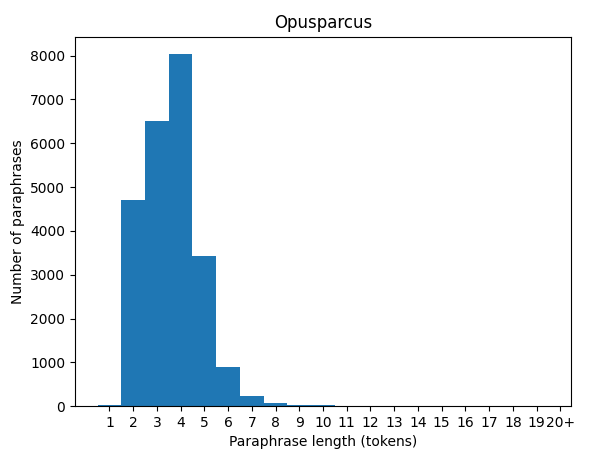}

\caption{Comparison of paraphrase length distributions in terms of tokens per paraphrase.}
\label{fig:length}
\end{figure*}

\begin{figure*}

\includegraphics[width=0.33\textwidth]{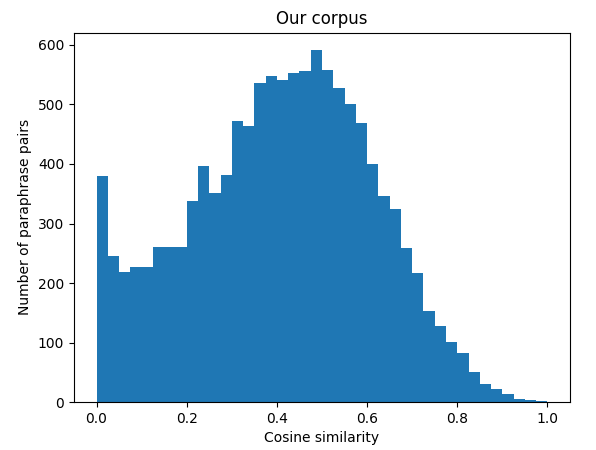}\hfill \includegraphics[width=0.33\textwidth]{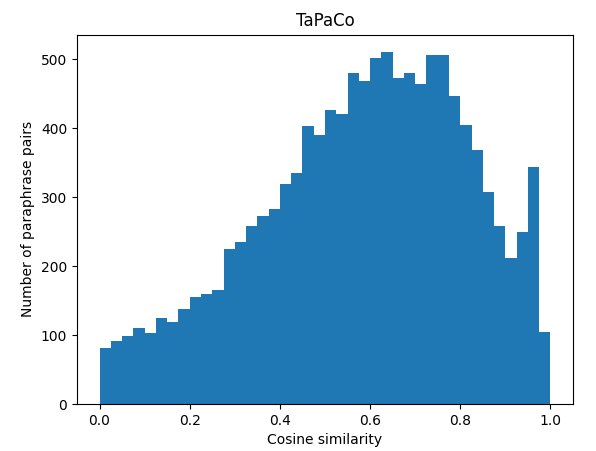}\hfill \includegraphics[width=0.33\textwidth]{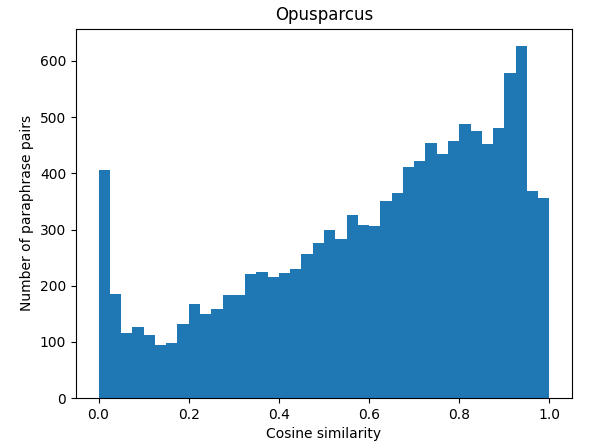}

\caption{Comparison of paraphrase pair cosine similarity distributions.}
\label{fig:sim-dist}
\end{figure*}

We compare the distribution of paraphrase lengths and lexical similarity with the two Finnish paraphrase candidate corpora, Opusparcus and TaPaCo, as the reference. Direct comparison is complicated by several factors. Firstly, both Opusparcus and TaPaCo consist primarily of automatically extracted paraphrase candidates, Opusparcus having only small manually curated development and test sections, and TaPaCo being fully uncurated. Secondly, the small manually annotated sections of Opusparcus are sampled to emphasize lexically dissimilar pairs, and therefore not representative of the characteristics of the whole corpus. We therefore compare with the fully automatically extracted sections of both Opusparcus and TaPaCo. For our corpus, we discard the small proportion of examples of base labels 1 and 2, i.e.\ not paraphrases. Another important factor to consider is that the proportion of false candidates in the automatically extracted sections of Opusparcus and TaPaCo is unknown, further decreasing comparability: the characteristics of false and true candidates may differ substantially, false candidates for example likely being on average more dissimilar in terms of lexical overlap than true candidates.

For each corpus, we sample 12,000 paraphrase pairs. For our corpus, we selected a random sample of true paraphrases (\emph{label 3} or higher) from the train section. For TaPaCo, the sample covers all paraphrase candidates from the corpus, however with the restriction of taking only one, random pair from each `set' of paraphrases. For Opusparcus, which is sorted by a confidence score in descending order, the sample was selected to contain the most confident 12K paraphrase candidates.\footnote{When we repeated the length analysis with a sample of 480K most confident pairs, the length distribution and average length remained largely unchanged, while the similarity distribution became close to flat. Without manual annotation, it is hard to tell the reason for this behavior.}

In Figure~\ref{fig:length} the length distribution of paraphrases in terms of tokens is measured for the abovementioned samples. Although the majority of paraphrases are rather short in all three corpora, we see that our corpus includes considerably higher proportion of longer paraphrases. The average number of tokens in our corpus is 8.3 tokens per paraphrase, while it is 5.6 in TaPaCo and 3.6 in Opusparcus candidates.

In Figure~\ref{fig:sim-dist} the paraphrase pair cosine similarity distribution is measured using TF-IDF weighted character n-grams of length 2--4. While both TaPaCo and Opusparcus lean towards higher similarity candidates, the distribution of our corpus is more balanced including a considerably higher proportion of pairs with low lexical similarity.

\section{Paraphrase Classification Baseline}

In order to establish a baseline classification performance on the new dataset, we train a classifier based on the FinBERT model \cite{virtanen2019multilingual}. Each paraphrase pair of statements \texttt{A} and \texttt{B} is encoded as the sequence \texttt{[CLS] A [SEP] B [SEP]}, where \texttt{[CLS]} and \texttt{[SEP]} are the special marker tokens of the BERT model. Subsequently, the output embeddings of the three special tokens are concatenated together with the averaged embeddings of the tokens in A and B. These five concatenated embeddings are then propagated into four decision layers: one for the base label \emph{2/3/4}, one for the subsumption flag \emph{$<$/$>$/none}, and one for each the binary flag \emph{s} and \emph{i}. Since the flags only apply to base label \emph{4}, no gradients are applied to these layers for examples with base labels \emph{2} and \emph{3}. We have explored also other BERT-based architectures, such as basing the classification on the \texttt{[CLS]} embedding only as is customary, and having a single classification layer comprising all possible base label and flag combinations. These resulted in a consistent drop in prediction accuracy, and we did not pursue them any further.

The baseline results are listed in Table~\ref{tab:classperf} showing that per-class F-score ranges between 38--71\%, strongly correlated with the number of examples available for each class. When interpreting the task as a pure multi-class classification, i.e.\ when counting all possible combinations of base label and flags as their own class, the accuracy is 54\% with majority baseline being 34.3\%, and the annotators' accuracy 68.7\%. The model thus positions roughly to the mid-point between the trivial majority baseline, and human performance. 

%\begin{tabular}{l|llll}
%    Label & Precision & Recall & F-score & Support\\
%    \hline
%    2             & 0.5741 & 0.3333 & 0.4218 & 93 \\
%    3             & 0.5214 & 0.4061 & 0.4566 & 990 \\
%    4             & 0.6789 & 0.7566 & 0.7157 & 2149 \\
%    4\textless    & 0.5478 & 0.5233 & 0.5353 & 1007 \\
%    4\textgreater & 0.5356 & 0.5625 & 0.5487 & 1136 \\
%    \hline
%    M. avg & 0.5716 & 0.5164 & 0.5356 & 5375 \\
%    W. avg & 0.5932 & 0.6000 & 0.5938 & \\
%    Acc & & & 0.6000 &
%\end{tabular}

\begin{table}[t]
\centering
\begin{tabular}{l|llll}
    \hline
    Label & Prec & Rec & F-score & Support\\
    \hline
    2             & 50.9 & 31.2 & 38.7 & 93 \\
    3             & 57.7 & 31.9 & 41.1 & 990 \\
    4             & 66.2 & 78.2 & 71.7 & 2149 \\
    4\textless    & 52.8 & 53.5 & 53.2 & 1007 \\
    4\textgreater & 52.6 & 56.1 & 54.3 & 1136 \\
    & & & & \\
    i             & 51.5 & 36.5 & 42.7 & 329 \\
    s             & 51.4 & 28.9 & 37.0 & 249 \\
    \hline
    W. avg & 52.9 & 54.0 & 52.2 \\
    Acc & & & 54.0 &
\end{tabular}
\caption{Classification performance on the test set, when the base label and the flags are predicted separately. In the upper section, we merge the subsumption flags with the base class prediction, but leave the \emph{i} and \emph{s} separated. The rows \emph{W. avg} and \emph{Acc} on the other hand refer to performance on the complete labels, comprising all allowed combinations of base label and flags. \emph{W. avg} is the average of P/R/F values across the classes, weighted by class support. \emph{Acc} is the accuracy.}
\label{tab:classperf}
\end{table}

\section{Discussion and Future Work}

In this work, we set out to build a paraphrase corpus for Finnish that would be (a) in the size category allowing deep model fine-tuning and (b) manually gathered maximizing the chance of finding more non-trivial, longer paraphrases than would be possible with the traditional automatic candidate extraction. The annotation so far took 14 person-months and resulted in little over 50,000 manually classified paraphrases. We have demonstrated that, indeed, the corpus has longer, more lexically dissimilar paraphrases. Building such a corpus is therefore shown feasible and presently it is likely the largest manually annotated paraphrase dataset for any language, naturally at the inevitably higher data collection cost. The manual selection is only feasible for texts rich in paraphrase, and the domains and genres covered by the corpus is necessarily restricted by this condition.

In our future work, we intend to extend the manually annotated corpus, ideally roughly double its present size. We expect the pursued data size will allow us to build sufficiently accurate models, both in terms of embedding and pair classification, to gather further candidates automatically at a level of accuracy sufficient to support down-stream applications. We are also investigating further text sources, especially parallel translations outside of the present subtitle domain. The additional flags in our annotation scheme, as well as the nearly 10,000 rewrites allow for interesting further investigations in their own right.

\section{Conclusions}

In this paper we presented the first entirely manually annotated paraphrase corpus for Finnish including 45,663 naturally occurring paraphrases gathered from alternative movie or TV episode subtitles and news headings. Further 7,909 hand-made rewrites are provided, turning context-dependent paraphrases into perfect paraphrases whenever possible. The total size of the released corpus is 53,572 paraphrase pairs of which 98\% are manually classified to be at least paraphrases in their given context if not in all contexts.

Additionally, we evaluated the advantages and costs of manual paraphrase candidate selection from two `parallel' but monolingual documents. We demonstrated the approach on alternative subtitles showing the technique being feasible for high quality candidate selection yielding sufficient amount of paraphrase candidates for the given annotation effort. We have shown the candidates to be notably longer and less lexically overlapping than what automated candidate selection permits.

The corpus is available at \url{github.com/TurkuNLP/Turku-paraphrase-corpus} under the CC-BY-SA license.

\section*{Acknowledgments}

We gratefully acknowledge the support of European Language Grid which funded the annotation work. Computational resources were provided by \emph{CSC --- the Finnish IT Center for Science} and the research was supported by the Academy of Finland. We also thank Sampo Pyysalo for fruitful discussions and feedback throughout the project.

%XXX TODO MUST NOT FORGET TO ACKNOWLEDGE ELG AND CSC!!!

%Do not number the acknowledgment section. Do not include this section
%when submitting your paper for review.

\bibliographystyle{acl_natbib}
\bibliography{nodalida2021}

\newpage
\onecolumn
\appendix\section{English Translation of Table~\ref{tab:annotations}}\label{sec:appendix}
\begin{table*}[h!]
\begin{tabular}{l|ll}
\hline
    Label  & Statement 1 & Statement 2  \\\hline
         4 & Shockingly childish! & Astoundingly immature! \\
         4s & I have worked for the duration of lunch. & I worked through the whole chowtime. \\
         4i & You guys got lucky, didn’t you. & Aren’t you fortunate. \\
         4$>$ & I worked so hard for the money. & I put so much effort into work. \\
         4$<$s & You rule! Come on, dude! & You are the best, Tähkä! Come on! \\
         4is & You nicked our plant! & You stole our plants! \\
         3 & I intend to make an experiment. & I am going to test it.  \\
         2 & Defeated Väyrynen vanished into the & Väyrynen is losing his seat in the parliament. \\
           & Helsinki night &   \\\hline
          & Rewrites & \\\hline
         Orig & Can I get back to my assignments?  & Can I continue?  \\
         Rew    & \emph{Can I get back to my assignments?} &\emph{Can I continue working on my assignments?} \\
        \hline
\end{tabular}
    \caption{English translations for annotation examples in Table~\ref{tab:annotations}.}
    \label{tab:annotations-eng}
\end{table*}

\end{document}